# R₂FD₂: Fast and Robust Matching of Multimodal Remote Sensing Image via Repeatable Feature Detector and Rotation-invariant Feature Descriptor

Bai Zhu[a, b], Chao Yang[a, b], Jinkun Dai[a, b], Jianwei Fan[c], Yao Qin[d], Yuanxin Ye[a, b, *]

[a]Faculty of Geosciences and Environmental Engineering, Southwest Jiaotong University, Chengdu 610031, China;

[b]State-province Joint Engineering Laboratory of Spatial Information Technology for High-speed Railway Safety, Southwest Jiaotong University, Chengdu 611756, China;

[c]School of Computer and Information Technology, Xinyang Normal University, Xinyang, 464000, China;

[d]Northwest Institute of Nuclear Technology, Xi'an 710025, China;

kevin_zhub@my.swjtu.edu.cn; yc18483685462@my.swjtu.edu.cn;

daijinkun@my.swjtu.edu.cn; fanjw@xynu.edu.cn; tsintuan@163.com.

* Corresponding author: Yuanxin Ye, yeyuanxin@home.swjtu.edu.cn.

**ABSTRACT:**

Automatically identifying feature correspondences between multimodal images is facing enormous challenges because of the significant differences both in radiation and geometry. To address these problems, we propose a novel feature matching method (named R₂FD₂) that is robust to radiation and rotation differences. Our R₂FD₂ is conducted in two critical contributions, consisting of a repeatable feature detector and a rotation-invariant feature descriptor. In the first stage, a repeatable feature detector called the Multi-channel Auto-correlation of the Log-Gabor (MALG) is presented for feature detection, which combines the multi-channel auto-correlation strategy with the Log-Gabor wavelets to detect interest points (IPs) with high repeatability and uniform distribution. In the second stage, a rotation-invariant feature descriptor is constructed, named the Rotation-invariant Maximum index map of the Log-Gabor (RMLG), which consists of two components: fast assignment of dominant orientation and construction of feature representation. In the process of fast assignment of dominant orientation, a Rotation-invariant Maximum Index Map (RMIM) is built to address rotation deformations. Then, the proposed RMLG incorporates the rotation-invariant RMIM with the spatial configuration of DAISY to depict a more discriminative feature representation, which improves RMLG's resistance to radiation and rotation variances. We conduct experiments to validate the matching performance of our R$_2$FD$_2$ utilizing three types of multimodal image datasets (optical-infrared, optical-LiDAR, and optical-SAR). Experimental results show that the proposed R$_2$FD$_2$ outperforms five state-of-the-art feature matching methods, and has superior advantages in adaptability and universality. Moreover, our R$_2$FD$_2$ achieves the accuracy of matching within two pixels and has a great advantage in matching efficiency over other state-of-the-art methods.

# 1. INTRODUCTION

With the launch of numerous multi-sensors integrated stereo observation facilities from spaceborne, airborne, and terrestrial platforms, a large number of multimodal remote sensing images (MRSIs) can be obtained at different times by different sensors (Yongjun et al., 2021). And the integration of MRSIs can provide complementary information for diverse applications in the field of image classification (Hao et al., 2017), bundle block adjustment (Cao et al., 2019), and change detection (Ye et al., 2022). There's a prerequisite for the integration of MRSIs, that is, the process of robust multimodal remote sensing image matching (MRSIM) is indispensable.

Generally, MRSIM aims to automatically identify accurate correspondences or control points (CPs) between two or more multimodal images (Zhu et al., 2022). However, there are extensive differences in scale, rotation, and radiation among MRSIs, and the image quality of different sensors is easily disturbed by noise, clouds, and blur, which are inevitable problems and entail enormous challenges for the reliable matching of MRSIs. Fig. 1 exemplarily shows these challenges of multimodal image matching mentioned above.

In view of the above challenges, numerous methods have been proposed for multimodal matching in the past two decades. These multimodal matching methods can be generally grouped into three categories: area-based methods, feature-based methods, and learning-based methods (Ma et al., 2021). With the development of deep learning technology, learning-based matching methods exhibit excellent matching performance and have developed into a pipeline that cannot be ignored in the field of MRSIM. Wang et al. (Wang et al., 2018) presented an effective deep neural network to optimize the whole processing (learning mapping function) through information feedback, and transfer learning was used to improve their framework's performance and efficiency. Contrapose different stages of feature matching, Hughes et al. (Hughes et al., 2020) proposed a fully-automated SAR-optical matching framework that is composed of a goodness network, correspondence network, and outlier reduction network, and each of these sub-networks has been proven to individually improve the matching performance. Further, Zhou et al. (Zhou et al., 2021) employed deep learning techniques to refine structure features and designed the multiscale convolutional gradient features (MCGFs) by utilizing a shallow pseudo-Siamese network. Similarly, Quan et al. (Quan et al., 2022) exploited more similar features using a self-distillation feature learning network (SDNet) for optimization enhancement of deep network, which achieves robust matching of optical-SAR images. Ye et al. (Ye et al., 2022) designed a multiscale framework without costly ground truth labels and a novel loss function paradigm based on structural similarity, which can directly learn the end-to-end mapping from multimodal image pairs to their transformation parameters. And their matching framework has the steady performance to be robust to NRD between multimodal image pairs.

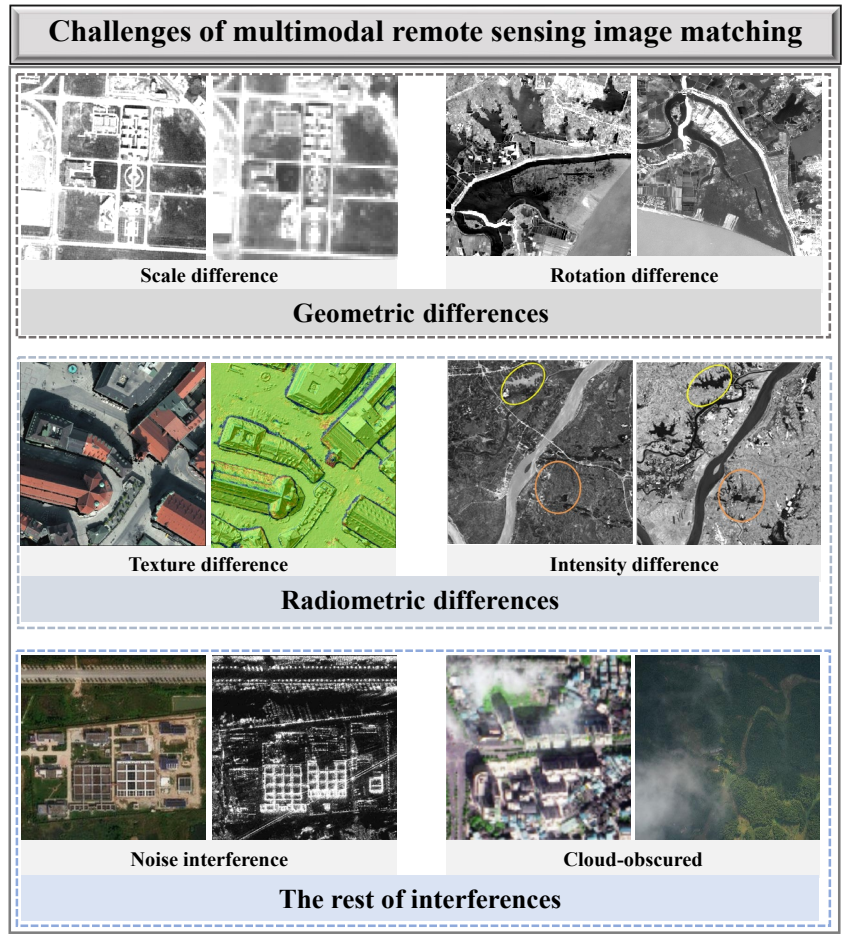

Fig. 1. Challenges of multimodal remote sensing image matching.

Although learning-based matching methods can significantly improve their resistance to geometric and radiation distortion by extracting finer common features than traditional handcrafted features, the main limitations of this pipeline are also found to be significant. On the one hand, supervised learning-based methods often rely on a large amount of training data (Hughes et al., 2020; Quan et al., 2022; Wang et al., 2018; Zhou et al., 2021), and the transferability of the trained model is poor results in their matching performance generally drops sharply on different test datasets. On the other hand, despite unsupervised learning methods that can overcome the dependence on training data, the process of converting various parameters is very complex (Ye et al., 2022), and its efficiency depends on the basic configuration of the computer infrastructure. These deficiencies mentioned above limit the widespread application of learning-based pipeline in multimodal matching fields.

Generally, traditional area-based methods identify correspondences by selecting some classical similarity metrics to evaluate the similarity of intensity information within a template window. There are three commonly used similarity metrics in the spatial domain: sum of squared differences (SSD), normalized cross-correlation (NCC), and mutual information (MI). In addition, phase correlation is the most commonly used similarity metric in the frequency domain because of its illumination invariance (Zhu et al., 2021). Recently, a structure feature-based pipeline has been developed for nonlinear radiometric differences (NRD) between multimodal images. These methods of this pipeline evaluate the similarity of generated features rather than intensity information by using the above similarity metrics (i.e., SSD, NCC, and phase correlation). Histogram of Orientated Phase Congruency (HOPC) (Ye et al., 2017), phase congruency structural descriptor (PCSD) (Fan et al., 2018), Channel Features of Orientated Gradients (CFOG) (Ye et al., 2019), and OS-PC (Xiang et al., 2020) are the most representative ones. However, area-based matching methods are very sensitive to geometric distortions (i.e., scale and rotation deformations) between images and usually require georeferencing implementation to eliminate the significant global geometric distortions (Ye et al., 2022).

In contrast, feature-based matching methods rely on the salient and distinctive features (i.e., points, lines, and regions) between images, and are more robust to geometric distortions and NRD compared with area-based methods (Sedaghat and Mohammadi, 2018). Among these methods, point features are the most common local invariant features in the remote sensing domain. This matching pipeline usually consists of two key components: feature detector and feature descriptor. In the past several decades, feature matching methods of monomodal images have been well-studied, and many classical feature detectors and feature descriptors have been developed. These traditional feature detectors detect salient features between images based on the gradient information of images, such as Moravec (Moravec, 1980), Harris (Harris and Stephens, 1988), differences of Gaussian (DoG) (Lowe, 2004), and Features from Accelerated Segment Test (FAST) (Rosten et al., 2008). Nevertheless, these gradient-based detectors are difficult to detect interest points (IPs) with high repeatability among multimodal images. According to the inherent properties of optical and SAR images, Xiang et al. (Xiang et al., 2018) constructed two Harris scale-spaces to extract IPs by designing consistent gradients for optical and SAR images utilizing the multiscale Sobel and multiscale ratio of exponentially weighted averages operators, respectively. Further, some studies have found that the use of the phase consistency (PC) model can effectively resistance to diverse significant NRD and extract more stable and repeatable IPs than using only the gradient information. Ye et al. (Ye et al., 2018) combined the minimum moment of PC with the Laplacian of Gaussian (MMPC-Lap) to detect stable IPs in image scale space. Subsequently, Li et al. (Li et al., 2019) detected corner feature points and edge feature points on the minimum moment map and maximum moment map of the PC, respectively. Although these PC-based feature detectors have a certain resistance to NRD between multi-modal images, they have the cost of high computational complexity.

Once the feature detection of MRSIs is completed, then corresponding local invariant feature descriptors must be explored. Similarly, the construction of many well-known feature descriptors also utilizes the gradient information of images, hence they cannot achieve the robust matching of MRSIs with both geometric distortion and radiation differences. Scale-Invariant Feature Transform (SIFT) (Lowe, 2004), Gradient Location and Orientation Histogram (GLOH) (Mikolajczyk and Schmid, 2005), DAISY (Tola et al., 2009), and their improved variants (Sedaghat and Ebadi, 2015; Sedaghat et al., 2011; Xiang et al., 2018) are the most representative feature descriptors. As shown in Fig 1, especially such significant intensity differences and severe speckle noise in multimodal images will further decline the matching performance of these gradient-based descriptors, accompanied by the difficulty of identifying accurate correspondences.

A recently popular pipeline for radiation-robust description is structural features because it is more resistant to modality variations than the gradient-based description that is already described in the above literature. With a number of descriptors derived from structural features having been developed for multimodal image matching, the most commonly used feature descriptors can be divided into two categories. The former is based on local self-similar (LSS) descriptors utilizing a Log-Polar spatial structure as feature descriptors, which can effectively capture the internal geometric composition of self-similarities within local image patches and is less sensitive to significant NRD to a certain extent (Sedaghat and Mohammadi, 2019; Xiong et al., 2021; Ye and Shan, 2014; Ye et al., 2017). Ye and Shan (Ye and Shan, 2014) introduced the LSS descriptor as a new similarity metric to detect correspondences for the matching of multispectral remote sensing images. Based on LSS, a shape descriptor named dense local self-similarity (DLSS) is further designed for optical and SAR image matching (Ye et al., 2017). Sedaghat and Mohammadi (Sedaghat and Mohammadi, 2019) improved the distinctiveness of the histogram of oriented self-similarity (HOSS) descriptor by adding directional attributes to image patches where the self-similarity values are computed. For the problem of LSS' computation complex, Xiong et al. (Xiong et al., 2021) proposed a feature descriptor named oriented self-similarity (OSS) by using the offset mean filtering to calculate the self-similarity features fast based on the symmetry of the self-similarity. However, there still exist limitations with these descriptors because the relatively low discriminative capability of LSS descriptors may lead to the inability to maintain robust matching performance in some multimodal matching cases (Fan et al., 2018).

Another structural feature of radiation-robust description is by utilizing the PC model, which is based on the position perception feature of the maximum Fourier component (Kovesi, 1999). Given that the PC model is more robust to illumination and contrast changes compared with gradient information, therefore, many PC-based descriptors have been developed (Fan et al., 2022; Li et al., 2019; Xiang et al., 2020; Ye et al., 2018). Ye et al. (Ye et al., 2018) presented a local histogram of orientated phase congruency (LHOPC) descriptor by combining the extended PC model and the arrangement of DAISY. Li et al. (Li et al., 2019) developed a radiation-variation insensitive feature transform (RIFT) method, and a maximum index map (MIM) was introduced based on the PC model for feature description. Xiang et al. (Xiang et al., 2020) improved different PC models to construct features for the matching of optical and SAR images. In a similar work, Fan et al. (Fan et al., 2022) designed a multiscale PC descriptor, named multiscale adaptive binning phase congruency (MABPC), that uses an adaptive binning spatial structure to encode multiscale phase congruency features, while improving its robustness to address geometric and radiometric discrepancies. Nevertheless, in the process of feature description, the above methods either lack rotation invariance (Xiang et al., 2020), or rely on time-consuming loop traversal based on the log-Gabor convolution sequence to achieve rotation-invariant (Li et al., 2019), or estimate the dominant orientation by combining orientation histogram with local PC features also is time-consuming and easily prone to generate outliers (Fan et al., 2022; Ye et al., 2018; Yu et al., 2021), which vastly affects the final matching performance.

Although numerous efforts have been made to enhance the robustness of MRSIM, the current feature detectors in the aspect of IP's repeatability are still not efficacious, and feature descriptors remain challenging in rotation invariance. To address the aforementioned limitations of pivotal components in feature matching, we present a robust and efficient feature-based method (called $R_2FD_2$) for multimodal matching in this work. First of all, to improve the repeatability of feature detection, we construct a repeatable feature detector called the Multi-channel Auto-correlation of the Log-Gabor (MALG). The MALG detector combines the multi-channel auto-correlation strategy with the Log-Gabor wavelets, which can be capable of extracting evenly distributed IPs with high repeatability. Subsequently, we build a rotation-invariant feature descriptor named the Rotation-invariant Maximum index map of the Log-Gabor (RMLG). The MALG descriptor consists of a fast assignment strategy of dominant orientation and an advanced descriptor configuration. In the process of fast assignment of dominant orientation, we propose a novel Rotation-invariant Maximum Index Map (RMIM) to achieve reliable rotation invariance. Then, the RMLG descriptor incorporates the rotation-invariant RMIM with the spatial configuration of DAISY to depict discriminative features of multimodal images, which aims to construct feature representation that is as robust as possible against the differences in radiation and rotation.

The following is a summary of the main contributions: (1) a repeatable feature detector called the MALG is defined to detect evenly distributed IPs with high repeatability. (2) a rotation-invariant feature descriptor named RMLG is constructed based on the RMIM with rotation invariance and the spatial configuration of DAISY. (3) the presented $R_2FD_2$ matching method, consisting of the MALG detector and RMLG descriptor, is quantificationally and qualitatively evaluated with existing state-of-the-art methods using various types of MRSIs.

The remainder of this article is organized as follows. The proposed multimodal feature matching method is introduced in Section II, with an emphasis on the construction of the MALG detector and RMLG descriptor. Section III examines and evaluates the matching performance of the proposed $R_2FD_2$ by conducting experiments on various multimodal image pairs. Finally, the conclusion is summarized in Section IV.

## 2. METHODOLOGY

In this section, a fast and robust method (named $R_2FD_2$), involving the MALG detector and the RMLG descriptor, is proposed to improve the matching performance of multimodal images. Specifically, the MALG detector is first presented to detect IPs with high repeatability between multimodal image pairs. Then, the RMLG descriptor is employed to robustly depict the local invariant characteristics of detected IPs. The flowchart of the proposed $R_2FD_2$ is illustrated in Fig. 2, which is then elaborated in detail.

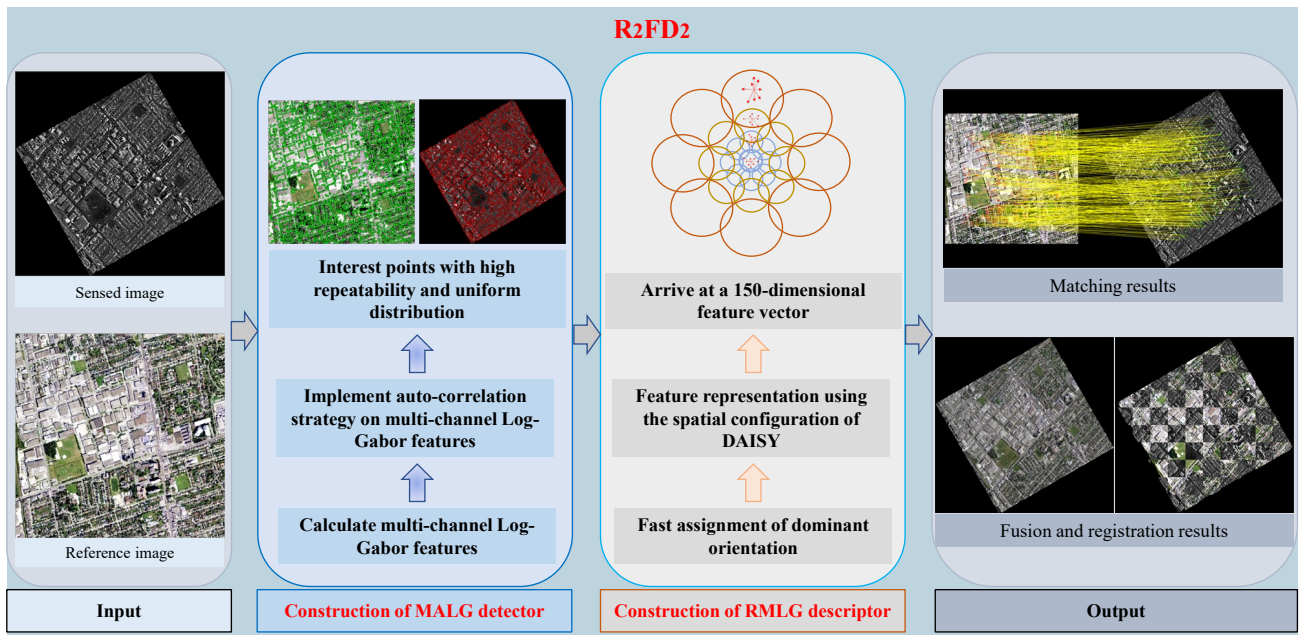

Fig. 2. Flowchart of the proposed $R_2FD_2$.

## 2.1 Construction of MALG detector

As mentioned above, the PC model is more resistant to significant NRD between multimodal images compared with gradient information, and there have been relevant studies (Li et al., 2019; Ye et al., 2018) using the PC model to extract stable IPs. Nevertheless, these detectors using the minimum moment or maximum moment of PC may cause loss of multi-directional features and are computationally expensive because they are the weighted responses of PC in different orientations and represent the moment changes with the orientation (Kovesi, 2003). And the responses of PC in different orientations are calculated by making use of Log-Gabor wavelets because of their good anti-noise and edge extraction performance (Xiang et al., 2017). In order to improve the reliability of detector while ensuring the high repeatability of IPs, in this paper, the MSLG detector is proposed by incorporating the multi-channel auto-correlation strategy with the Log-Gabor wavelets for IPs detection.

Given good noise suppression and edge preservation are two crucial characteristics of an excellent feature detector (Fan et al., 2014), the 2D Log-Gabor wavelets are employed during the construction of the proposed MALG detector. They can provide a useful description of edge feature information for multiple orientations and at multiple scales from multimodal image pairs, which is suitable for describing the local structure of multimodal images. Generally, a 2D Log-Gabor filter is expressed as follows:

$$LG_{s,o}(f,\theta) = \exp\left\{-\frac{(\log(f/F_s))^2}{2(\log\beta)^2}\right\}\exp\left\{-\frac{(\theta-\theta_o)^2}{2\delta_\theta^2}\right\} \qquad (1)$$

Where $o$ and $s$ represent the orientation and scale of the Log-Gabor filter, respectively. $\beta$ determines the bandwidth of the filter. $f$ and $F_s$ defines the frequency and central frequency of the filter, respectively. And $\delta_\theta$ is the angular bandwidth, $\theta_o$ represents the filter's orientation.

Since the 2D log-Gabor is a frequency domain filter, its expression in the space domain can be obtained by inverse Fourier transform based on the corresponding frequency response of log-Gabor filters in polar coordinates (Arrospide and Salgado, 2013). Therefore, the 2D log-Gabor function in the space domain can be typically decomposed into an even-symmetric filter and an odd-symmetric filter, which is defined as follows:

$$LG(x,y,s,o) = LG^{even}(x,y,s,o) + i \cdot LG^{odd}(x,y,s,o) \qquad (2)$$

Where the real component $LG^{even}(x,y,s,o)$ and the imaginary component $LG^{even}(x,y,s,o)$ represent the even-symmetric and the odd-symmetric filters, respectively, of the log-Gabor wavelets at scale $s$ with orientation $o$.

Accordingly, the space response components $E(x,y,s,o)$ and $O(x,y,s,o)$ of log-Gabor filters can be yielded by convolving the image $I(x,y)$ with the two even- and odd-symmetric filters:

$$\begin{cases} E(x,y,s,o) = I(x,y) * LG^{even}(x,y,s,o) \\ O(x,y,s,o) = I(x,y) * LG^{odd}(x,y,s,o) \end{cases} \quad (3)$$

Then, the amplitudes of Log-Gabor for all $Ns$ scales are summed at orientation $o$ to obtain the multi-channel Log-Gabor features, formally, the Multi-channel Log-Gabor features are defined as Eq. (4).

$$\begin{cases} A(x,y,s,o) = \sqrt{E_{s,o}^2(x,y) + O_{s,o}^2(x,y)} \\ A^o = \left[A_i(x,y,o)\right]_1^{N_o} = \sum_1^{N_s} A(x,y,s,o) \end{cases} \quad (4)$$

Where $A(x,y,s,o)$ is the amplitude component of $I(x,y)$ at scale $s$ and orientation $o$. $Ns$ represents the number of scales for the Log-Gabor filter banks, $No$ represents the number of orientations for the Log-Gabor filter banks, and $\Sigma$ is summed over the Log-Gabor filter banks on different scales for the orientation $o$. And $A_i(x,y,o)$ equals the amplitude responses of the Log-Gabor at the location $(x,y)$ for orientation $o$, and $1 \leq i \leq No$. In this paper, $Ns=4$ and $No=6$ are fixed values.

For the multi-channel Log-Gabor features $A_i(x,y,o)$, their self-similarity for Log-Gabor features of each orientation after a shift $(\Delta x, \Delta y)$ at the location $(x,y)$ can be yielded by the following auto-correlation function:

$$CA^o = C\left[A_i(x,y,\Delta x,\Delta y,o)\right]_1^{N_o} = \sum_{(u,v)\in W(x,y)} w(u,v)\left[A^o(x,y,o) - A^o(u+\Delta x, v+\Delta y, o)\right]^2 \quad (5)$$

Where $W(x,y)$ is a window centered at the location $(x,y)$. And $w(u,v)$ is a weighting function, which is either a constant or a Gaussian weighting function. According to Taylor expansion, the first-order approximation is performed after shifting $(\Delta x, \Delta y)$ for the Log-Gabor feature of each channel:

$$\begin{aligned} A^o(u+\Delta x, v+\Delta y, o) &= A^o(u,v,o) + A_x^o(u,v,o)\Delta x + A_y^o(u,v,o)\Delta y + O(\Delta x^2 + \Delta y^2) \\ A^o(u+\Delta x, v+\Delta y, o) &\approx A^o(u,v,o) + A_x^o(u,v,o)\Delta x + A_y^o(u,v,o)\Delta y \end{aligned} \quad (6)$$

Where $A_x^o$ and $A_y^o$ are the partial derivative of Log-Gabor feature in corresponding orientation $o$. Therefore, the above autocorrelation function for each orientation can be simplified as:

$$CA^o = C\left[A_i(x,y,\Delta x,\Delta y,o)\right]_1^{N_o} = |\Delta x, \Delta y| M(x,y,o)[\Delta x, \Delta y] \tag{7}$$

with $M(x,y,o)$ denoting the auto-correlation matrix of orientation $o$ defined as:

$$M(x,y,o) = \begin{bmatrix} \sum_w A_x^o(x,y,o)^2 & \sum_w A_x^o(x,y,o)A_y^o(x,y,o) \\ \sum_w A_x^o(x,y,o)A_y^o(x,y,o) & \sum_w A_y^o(x,y,o)^2 \end{bmatrix} \tag{8}$$

Then, autocorrelation features in all orientations are combined to obtain a comprehensive autocorrelation matrix (denoted as $M_{com}$) with multi-directional features:

$$M_{com}(x,y) = \begin{bmatrix} \sum_1^{N_o}\left[\sum_w A_x^o(x,y,o)^2\right] & \sum_1^{N_o}\left[\sum_w A_x^o(x,y,o)A_y^o(x,y,o)\right] \\ \sum_1^{N_o}\left[\sum_w A_x^o(x,y,o)A_y^o(x,y,o)\right] & \sum_1^{N_o}\left[\sum_w A_y^o(x,y,o)^2\right] \end{bmatrix} \tag{9}$$

The autocorrelation response value $R$ of the multi-channel Log-Gabor features for each pixel is calculated by utilizing the comprehensive autocorrelation matrix $M_{com}$:

$$R = \det[M_{com}(x,y)] - \alpha[traceM_{com}(x,y)]^2 \tag{10}$$

Where $\det[M_{com}(x,y)]$ is the determinant of matrix $M$, $traceM_{com}(x,y)$ represents the direct trace of matrix $M_{com}$. And α is a constant, ranging from 0.04 to 0.06. Finally, the local maximum extrema are first the extracted IP, while non-maximum suppression is carried out to decrease some adjacent IPs. That is, the first N local extrema with the large response values will be selected as the final IPs by our MALG detector. And the implementation of MALG is summarized in Alg. 1.

| Algorithm 1. The implementation of MALG |
|---|
| **Input:** A multimodal image pairs |
| **Start:** |
|    1: Initialize parameters (*Ns*=4 and *No*=6); |
|    2: Calculate the multi-channel Log-Gabor features $A^o$ using Eqs. (1), (2), (3), and (4); |
|    3: Calculate the self-similarity for Log-Gabor features of each orientation using the auto-correlation function (5) and (6); |
|    4: Calculate the auto-correlation matrix for each orientation $M(x,y,o)$ using the simplified auto-correlation function (7) and (8); |
|    5: Obtain the comprehensive autocorrelation matrix $M_{com}$ using Eq. (9); |
|    6: Calculate the autocorrelation response value *R* of the multi-channel Log-Gabor features for each pixel using Eq. (10); |
|    7: Perform non-maximum suppression; |
|    8: Select the first N local extrema with the large response values as the final interest points (IPs). |
| **End** |
| **Output:** Interest points (IPs) with high repeatability and uniform distribution |

Moreover, Fig. 3 presents three illustrations of the IPs extracted by the proposed MALG detector, specifically, Fig. 3(a) to Fig. 3(c) are the IPs extracted from optical-infrared, optical-depth, and optical-SAR image pairs respectively. As seen, our MALG detector is capable of extracting IPs with high repeatability and uniform distribution between multimodal image pairs. The definition of repeatability is introduced in Section 3.2, and more performance evaluation of MALG is given in a later Section 3.2.

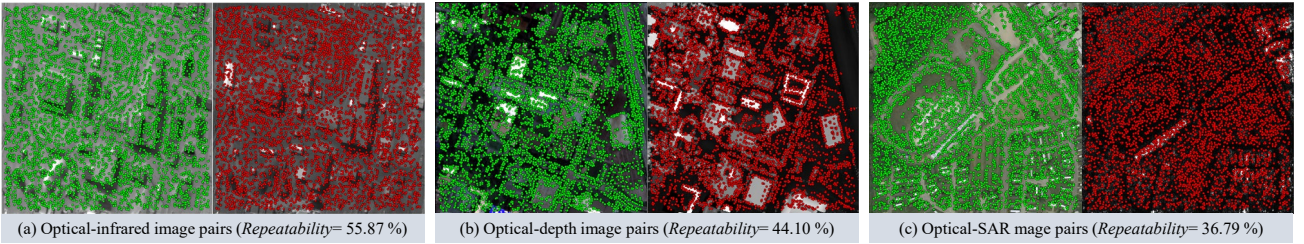

(a) Optical-infrared image pairs (*Repeatability*= 55.87 %)   (b) Optical-depth image pairs (*Repeatability*= 44.10 %)   (c) Optical-SAR mage pairs (*Repeatability*= 36.79 %)

Fig. 3. Schematic illustration of extracted IPs by the proposed MALG detector.

**2.2 Establishment of RMLG descriptor**

Once repeatable IPs are extracted, the next critical step is to design a robust feature descriptor with the intent of increasing the distinction of features. The feature descriptor usually consists of two components: assignment of dominant orientation and construction of feature representation. However, as mentioned earlier, the gradient-based descriptors are very sensitive to NRD, and existing structural feature-based descriptors either rely on time-consuming loop traversal based on the log-Gabor convolution sequence to achieve rotation-invariant (Li et al., 2019), or complicatedly assign the dominant orientation combining orientation histogram with local PC features (Fan et al., 2022; Ye et al., 2018; Yu et al., 2021). Therefore, it is difficult for these descriptors to achieve fast and robust multimodal image matching. To overcome these problems, we first propose a fast strategy for assigning dominant orientation and further employ the spatial configuration of DAISY for feature representation, and the above two components are combined to generate the final RMLG descriptor. More details regarding the proposed RMLG descriptor are provided below.

**2.2.1   Fast assignment of dominant orientation**

From the previous Eq. (4) we can get the Multi-channel Log-Gabor features $A^o$, that is, the log-Gabor response sequence $[A_i(x,y,o)]_1^{N_o}$. Nevertheless, the log-Gabor response sequence not possesses the rotation invariance compared with the gradient map. This means that rotating the log-Gabor response sequence will not yield the corresponding log-Gabor response sequence for the rotated image patch. To obtain the rotation invariance, the Maximum Index Map (MIM) and circular effect are proposed by means of loop traversal (Li et al., 2019). The calculation of MIM is as follows:

$$MIM(x,y) = \arg\max_{o} \left\{ [A_i(x,y,o)]_1^{N_o} \right\} \tag{11}$$

Where $\arg\max_{o}$ represents the orientation index corresponding to the maximum value in the log-Gabor response sequence $[A_i(x,y,o)]_1^{N_o}$.

Further, Yu et al. (Yu et al., 2021) assigned the dominant orientation by combining the orientation histogram with the amplitudes and orientations of PC, and auxiliary orientations were also estimated in the same way as the SIFT. Then, the corresponding MIM patch was rotated by the dominant or auxiliary orientations, subsequently, the index of the rotated MIM patch for the reference and sensed image was cyclically shifted by $k_{ref}$ and $k_{sen}$ positions, respectively.

$$k_{ref} = round(\frac{rotation}{180° / N_o}) \tag{12}$$

$$k_{sen} = \begin{cases} ceil(\frac{rotation}{180° / N_o}) \\ floor(\frac{rotation}{180° / N_o}) \end{cases} \tag{13}$$

Where *round* indicates the rounding operation. And *ceil* and *floor* represent the round-up and round-down operation, respectively.

That is, one feature vector was constructed for each orientation of IPs in the reference image, and two feature vectors are constructed for each orientation of IPs in the sensed image. The aforementioned loop traversal strategy to achieve rotation invariance is very time-consuming. While Yu et al. 's (Yu et al., 2021) estimation of the dominant orientation requires calculating the complex amplitudes and orientations of PC, and increasing the auxiliary directions of feature points, and designing two feature vectors for each orientation of IPs in the sensed image, which will further lead to the time-consuming nature of their descriptor.

We note that the essence of the orientation histogram in the SIFT descriptor is to count the gradient amplitude and orientation of the pixels in the neighborhood, and the orientation corresponding to the peak of the histogram represents the dominant directions of IPs. The range of the gradient orientation is [0, 360], and it's continuous, while the value range of MIM based on the log-Gabor response sequence is [1, *No*], it's very discrete. Therefore, there are many redundant orientation estimations based on the orientation histogram, because the index of the rotated MIM patch needs to be cyclically shifted by *kref* and *ksen* positions in the reconstruction of MIM.

Inspired by the orientation histogram of SIFT and combined with the above analysis, we design a fast assignment strategy of dominant orientation, and a novel MIM with rotation invariance is performed by a statistical measure based on the MIM. This strategy can avoid the process of weighting the histogram calculation by trilinear interpolation that calculates the weight of each pixel of the spatial and directional bins. Specifically, the fast assignment strategy of dominant orientation is calculated as follows.

The essence of the orientation histogram is to count the gradient amplitude and orientation of the pixels in the neighborhood, while MIM itself has the directional characteristics of the Log-Gabor convolution sequence. Hence, we directly count the value with the most occurrences in the MIM (denoted as $C_{MIM}$), and use it as the dominant orientation of IPs, which can be expressed as follows:

$$\begin{cases} C_{MIM} = \mathrm{mode}[MIM(x,y)] \\ DO = C_{MIM} * \dfrac{180°}{N_o} \end{cases} \quad (14)$$

Where mode represents the operation to calculate the sample mode in MIM, that is, the value that appears most times in MIM. And *DO* represents the dominant orientation. Fig. 4 illustrates the feasibility of the above strategy for calculating dominant orientation with different rotated images. Given a reference image without rotation, its corresponding sensed image without rotation, and rotating 90° the sensed image, we select a pair of corresponding IPs between these images, then their dominant orientations are computed respectively. It is not difficult to find that the dominant orientation of the proposed strategy is the same, and this example preliminarily indicates that the proposed strategy of dominant orientation is feasible.

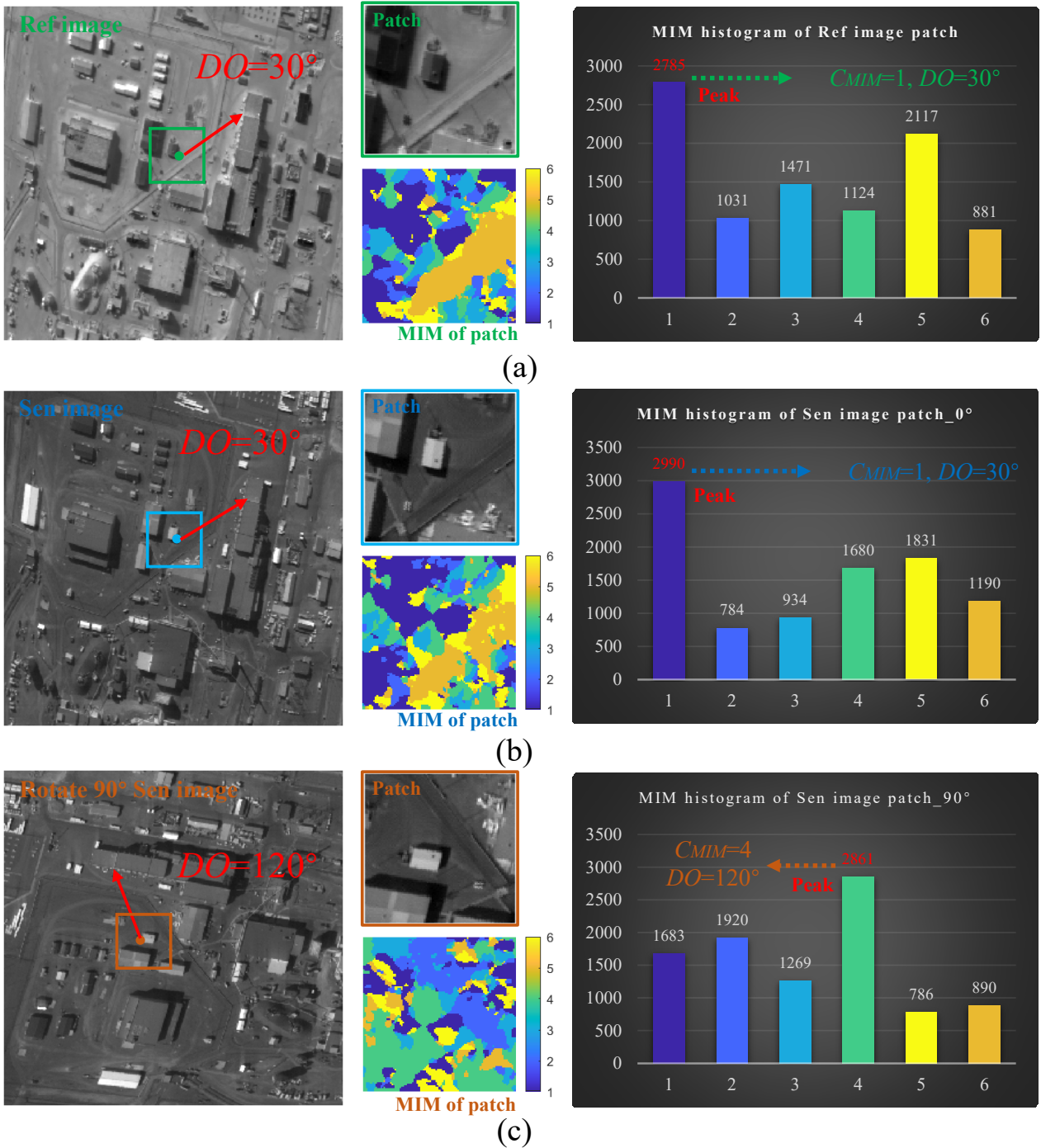

Fig. 4. Schematic illustration of the proposed strategy for calculating dominant orientation. The bar charts represent the MIM histograms of the local regions around the IPs. (a) is the dominant orientation of the reference image patch without rotation. (b) is the dominant orientation of the sensed image patch without rotation. (c) is the dominant orientation of the rotated sensed image patch with 90◦ rotation.

What follows is the reconstruction of MIM, $C_{MIM}$ is used to calculate the new MIM based on the following equation:

$$\begin{cases} MIM_{new}(x,y) = MIM(x,y) - C_{MIM} + 1 \\ MIM_{new}(x,y) = MIM_{new}(x,y) + N_o, MIM_{new}(x,y) < 1 \end{cases} \quad (15)$$

Actually, $MIM_{new}(x,y)$ represents the new MIM that is recalculated by circularly shifting the C*MIM*-th layer of the Log-Gabor convolution sequence as the first layer of the Log-Gabor convolution sequence. Finally, the novel MIM with rotation invariance (named RMIM) can be obtained by rotating the recalculated MIM by the dominant orientation:

$$RMIM = rotate[MIM_{new}(x,y), DO] \quad (16)$$

To verify the rotation invariance of the proposed RMIM more intuitively, we perform a contrast experiment to compare the rotation invariance of the MIM and RMIM, as shown in Fig. 5. Fig. 5(f) is obtained by rotating Fig. 5(a) anticlockwise 30°; Fig. 5 (b) and Fig. 5 (g) are the MIM of Fig. 5 (a) and Fig. 5 (f), respectively; Fig. 5 (c) and Fig. 5 (h) are the $MIM_{new}$ of Fig. 5 (a) and Fig. 5 (f), respectively; Fig. 5 (d) and Fig. 5(i) are the RMIM of Fig. 5 (a) and Fig. 5 (f), respectively. Fig. 5 (e) is the difference in MIM between Fig. 5 (b) and Fig. 5 (g); Fig. 5 (j) is the difference in RMIM between Fig. 5 (d) and Fig. 5 (i). There are significant differences in MIM between Fig. 5 (b) and Fig. 5 (g), and most of the values of Fig. 5 (e) are not close to zero. Nevertheless, most of the values of Fig. 4(j) are close to zero, it is not difficult to find that the similarity of RMIM between Fig. 5 (d) and Fig. 5 (i) is obvious. This substantially indicates that the generated RMIM based on our proposed fast assignment strategy of dominant orientation is rotationally invariant. And more evaluation of RMLG's rotation invariance is proved in Section 3.3.

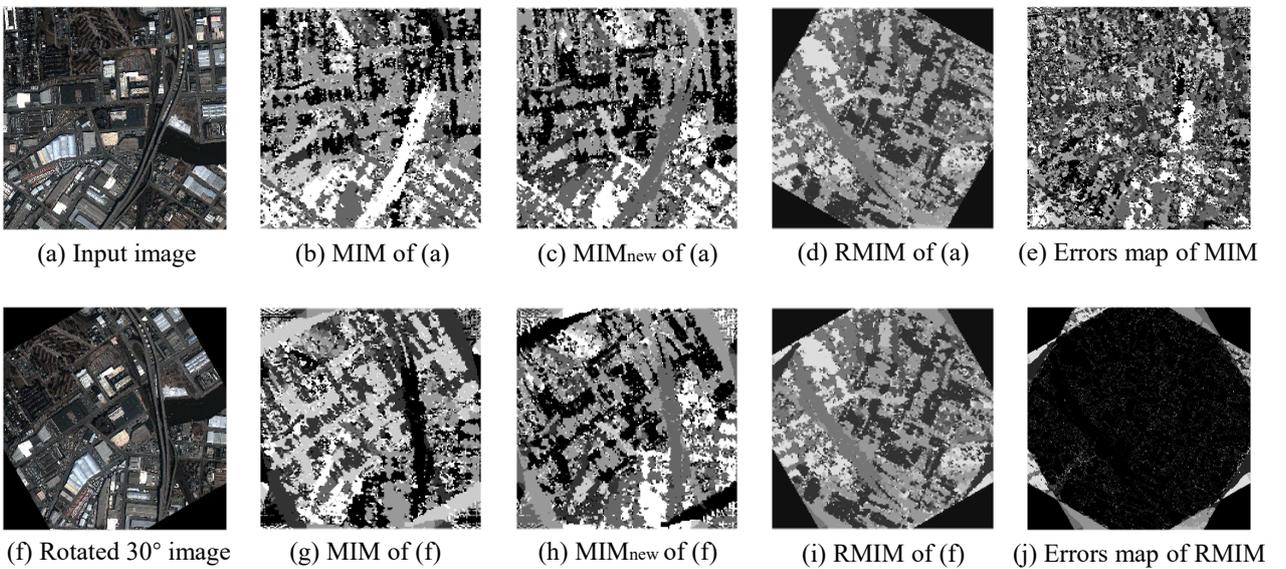

(a) Input image　　(b) MIM of (a)　　(c) MIMnew of (a)　　(d) RMIM of (a)　　(e) Errors map of MIM

(f) Rotated 30° image　　(g) MIM of (f)　　(h) MIMnew of (f)　　(i) RMIM of (f)　　(j) Errors map of RMIM

Fig. 5. Comparison of rotation invariance for the MIM and RMIM.

**2.2.2 RMLG descriptor representation**

After building the RMIM with rotation invariance, the follow-up critical step is to generate a unique feature description utilizing the RMIM. In the construction of feature descriptors, a reasonable spatial arrangement for feature description is crucial. Different spatial arrangements for feature description have been proposed, the most representative of which are the square grid in SIFT (Lowe, 2004), the log-polar grid in GLOH (Mikolajczyk and Schmid, 2005), and the circular grid in DAISY (Tola et al., 2009). Fig. 6 shows these different spatial arrangements.

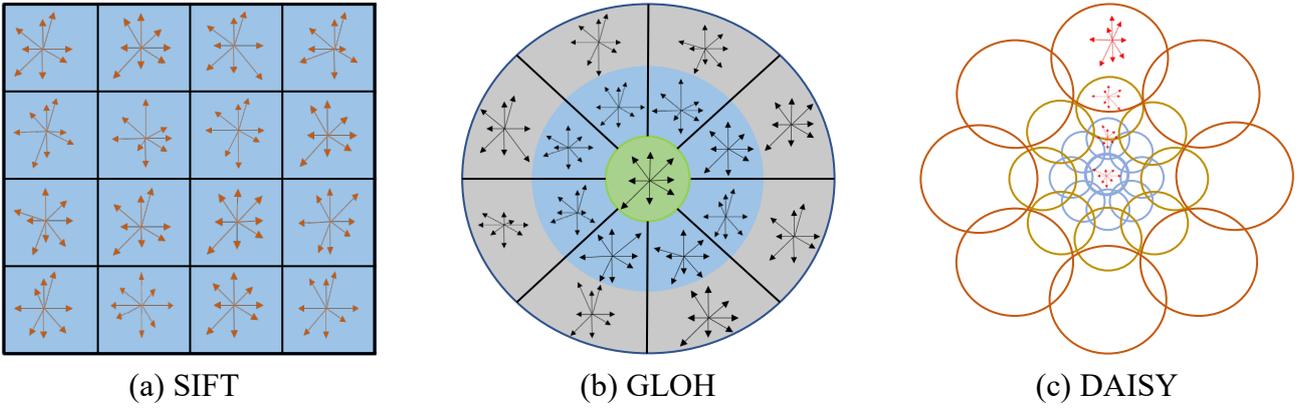

(a) SIFT  (b) GLOH  (c) DAISY

Fig. 6. Three different spatial arrangements for feature description.

Moreover, combining Fig. 5(d), Fig. 5(i), and Fig. 5(j), we can easily find that the rotation invariance of the proposed RMIM is extremely weak in the edge region, and its rotation invariance mainly exists in the circular neighborhood near the center point of the RMIM map. Therefore, it may be more effective to use a DAISY-style spatial arrangement (i.e., circular grid) for the construction of our RMLG descriptor based on a distribution histogram technique. Due to the values of RMIM ranging from 1 to *No* (*No*=6), and the final feature vector is obtained by concatenating all the histograms, therefore, the dimension of the feature vector is Bins × *No*.

To further verify the above conjecture, we use the aforementioned spatial arrangements (i.e., SIFT-style, GLOH-style, and DAISY-style) to construct the RMLG descriptor with different spatial arrangements on the RMIM. Since the dimensions of the feature vectors constructed by the different spatial arrangements are also different, in order to make a fair comparison with the spatial arrangement of DAISY, we have improved the spatial arrangement of SIFT and GLOH, so that the dimensions of all generated feature vectors are about 150.

Specifically, the initial and improved spatial arrangement of SIFT is the 4*4 and 5*5 square grid, so a 96- and 150-dimensional feature vector can be obtained utilizing the values of RMIM ranging from 1 to *No* (*No*=6), respectively. While the improved spatial arrangement of GLOH is a reference to the configuration of Yu et al (Yu et al., 2021), then a 144-dimensional feature vector is obtained. And the RMLG descriptor by combining the RMIM with the arrangement of DAISY is a 150-dimensional feature vector. We conduct a comparison experiment of the RMLG descriptor with different spatial arrangements on three different multimodal datasets (optical-infrared, optical-LiDAR, optical-SAR) for a total of sixty image pairs (More details of these datasets are introduced in Section 3.1). Table 1 gives the average Number of Correct Matches (NCM) of the RMLG descriptor with different spatial arrangements. It is obvious that the RMLG descriptor with the spatial arrangement of DAISY yields the best matching performance.

Table 1. Average matching performance of the RMLG descriptor with different spatial arrangements.

| Different spatial arrangements of RMLG | | Bins | Descriptor dimension | Average NCM (Optical-infrared) | Average NCM (Optical-LiDAR) | Average NCM (Optical-SAR) |
|---|---|---|---|---|---|---|
| SIFT-style | Initial | 4*4=16 | 16*6=96 | 323 | 272 | 143 |
| | Improved | 5*5=25 | 25*6=150 | 388 | 312 | 231 |
| GLOG-style | Initial | 2*8+1=17 | 17*6=102 | 355 | 297 | 177 |
| | Improved | 3*8=24 | 24*6=144 | 401 | 324 | 246 |
| DAISY-style | initial | 3*8+1 =25 | 25*6=150 | **448** | **355** | **272** |

As a consequence, the proposed RMLG is finally constructed by applying the arrangement of DAISY, arriving at a 150-dimensional feature vector, because the spatial arrangement of DAISY has been shown to outperform other spatial arrangements (e.g., SIFT and GLOH). And a more detailed analysis of RMLG's matching performance will be presented in Section 3.2.

## 2.3 $R_2FD_2$ Feature matching

In this paper, the proposed $R_2FD_2$ matching method is composed of the constructed MALG detector and RMLG descriptor. First, IPs with high repeatability are detected from a reference image and a sensed image by the proposed MALG detector, then their local invariant features are calculated utilizing the proposed RMLG descriptor. Finally, the commonly used Nearest Neighbor Distance Ratio (NNDR) (Lowe, 2004) matching strategy is employed to identify initial correspondences between reference and sensed image pairs, and the Fast Sample Consensus (FSC) (Wu et al., 2014) technique is performed to remove outliers for determining reliable correspondences.

## 3. EXPERIMENTAL EVALUATION AND ANALYSIS

In this section, to validate the matching performance of the proposed $R_2FD_2$ matching method, we selected different types of MRSIs datasets (e.g., optical-infrared, optical-LiDAR, and optical-SAR image pairs) for qualitative and quantitative evaluation. We first introduced the experimental settings, including the parameters predefined and the details of MRSIs datasets. Next, we evaluated the repeatability performance of MALG detector and the rotation invariance of RMLG descriptor, respectively. Finally, the matching results of $R_2FD_2$ were presented and analyzed by comparing them with the five state-of-the-art matching methods (including OS-SIFT (Xiang et al., 2018), HOSS (Sedaghat and Mohammadi, 2019), RIFT (Li et al., 2019), RI-ALGH (Yu et al., 2021), and MS-HLMO (Gao et al., 2022)) for verifying the robustness and effectiveness of $R_2FD_2$.

### 3.1 Experimental settings

We collect three types of MRSIs datasets for qualitative and quantitative evaluations, including optical-infrared, optical-LiDAR, and optical-SAR datasets. Each type of dataset consists of twenty image pairs for a total of sixty multimodal image pairs. The size of image pairs ranges from 450 × 450 pixels to 750 × 750 pixels. These experimental multimodal datasets include various high, medium, and low spatial resolution images, covering both urban and suburban areas, and there are significant radiation differences between multimodal image pairs. These challenges can comprehensively test the robustness and adaptability of the proposed matching method. Several sample image pairs of each type of dataset are shown in Fig. 7. It is worth noting that since our $R_2FD_2$ is not currently scale-invariant, the resolution of each image pair is resampled with the same ground sample distance (GSD).

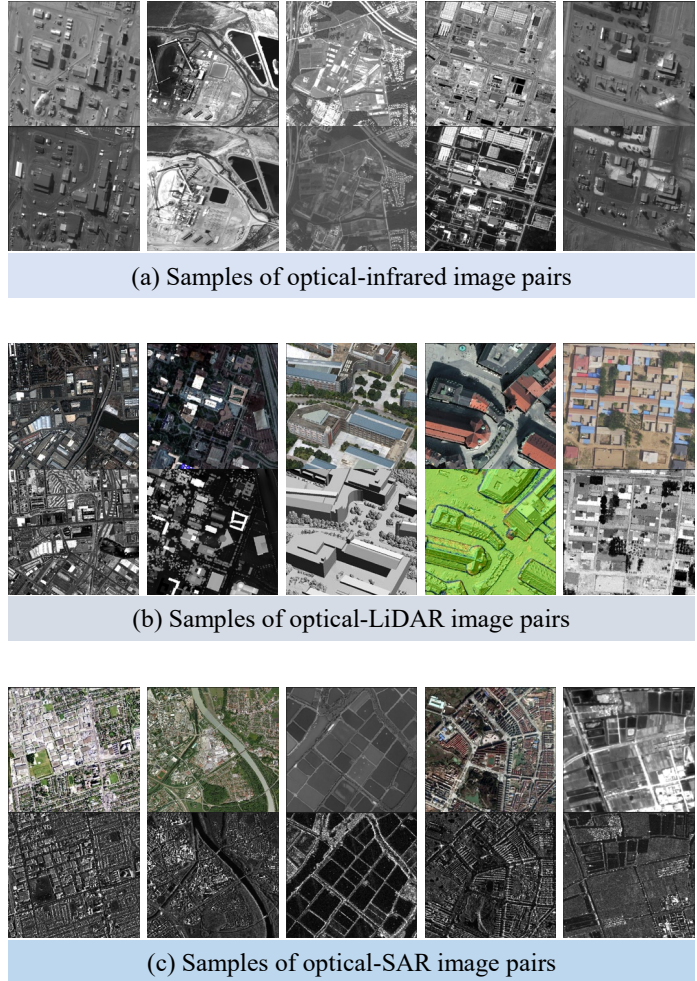

(a) Samples of optical-infrared image pairs

(b) Samples of optical-LiDAR image pairs

(c) Samples of optical-SAR image pairs

Fig. 7. Samples of MRSIs datasets.

As aforementioned in Section 2, the number of scales and orientations (i.e., $N_s$ and $N_o$) for the Log-Gabor wavelets were set to 4 and 6, respectively. And the RMLG descriptor of our $R_2FD_2$ was implemented by employing the arrangement of DAISY, therefore, the local region of feature description consists of many centrosymmetric circles of different sizes. These circles are located on a concentric structure of three layers with different radii, and eight circles are uniformly distributed in each layer, which finally generates our $R_2FD_2$ with 150-dimensional features.

The codes of OS-SIFT, HOSS, and MS-HLMO were provided by the corresponding author's website. Since the code exposed by RIFT is not rotationally invariant, we reproduce it based on the relevant content of RIFT's paper. Similarly, we reproduce RI-ALGH's rotation invariance (excluding the scale invariance component) based on the description of the paper for comparison because it not has public code, and our MALG detector was used for IP extraction before RI-ALGH feature description in subsequent experiments. The parameters of all comparison methods are set to the optimal settings recommended in the corresponding papers. All experiments were implemented by using MATLAB2020a on a personal computer with the configuration of Inter (R) Core (TM) CPU i7-10750H 2.6 GHz and 16 GB RAM.

### 3.2 Repeatability evaluation of MALG detector

To verify the repeatability performance of our proposed MALG on feature detection, we conducted a comparison experiment of IPs' repeatability on aforesaid multimodal datasets (e.g, optical-infrared, optical-LiDAR, optical-SAR) for a total of sixty image pairs. In the process of IPs detection, the repeatability of IPs is a very important index. Generally speaking, the higher the repeatability rate of IPs between two images, the more IPs can potentially be matched as correspondences (Mikolajczyk and Schmid, 2005). The calculation of IPs' repeatability is represented by the following equation:

$$Repeatability = \frac{N_{cor}}{0.5*(N_{ref} + N_{sen})} \tag{17}$$

Where $N_{ref}$ and $N_{sen}$ represent the number of IPs detected from the reference image and sensed image, respectively. $N_{cor}$ represents the number of correspondences whose location error (denoted as $E_{loc}$) smaller than a certain threshold (3 pixels) through the prior mapping relationship between the reference and sensed images, which can be expressed as follows:

$$E_{loc} = |ref(x,y) - P_{truth} * sen(x,y)| \tag{18}$$

Where $ref(x,y)$ and $sen(x,y)$ represent the extracted IPs on the reference and sensed images, respectively. $P_{truth}$ is the projective model and can be obtained by using manually selected correspondences.

Table 2. Comparison results of IPs' repeatability.

| Category | Average repeatability (%) | |
|---|---|---|
| | MALG | FAST$_{MPC}$ |
| Twenty optical-infrared pairs | 57.38 | 51.50 |
| Twenty optical-LiDAR pairs | 40.32 | 35.42 |
| Twenty optical-SAR pairs | 34.14 | 27.84 |

Specifically, 5000 IPs were obtained by our MALG detector, and FAST detector on the maximum and minimum moment of PC (denoted as $FAST_{MPC}$), respectively. Table 2 gives the average repeatability rate of extracted IPs by the two detectors on different types of multimodal images. Fig. 8 shows the comparison results of the repeatability and distributions of IPs between the $FAST_{MPC}$ and our MALG detector, and a representative example of each type of dataset is presented.

It is not difficult to find that the average repeatability rate of IPs by our MALG detector can be increased by about five percentage points than the $FAST_{MPC}$ detector. What's more, the MALG detector can extract more evenly distributed IPs in multimodal image pairs compared with the $FAST_{MPC}$ detector which was prone to the phenomenon of IPs gathering.

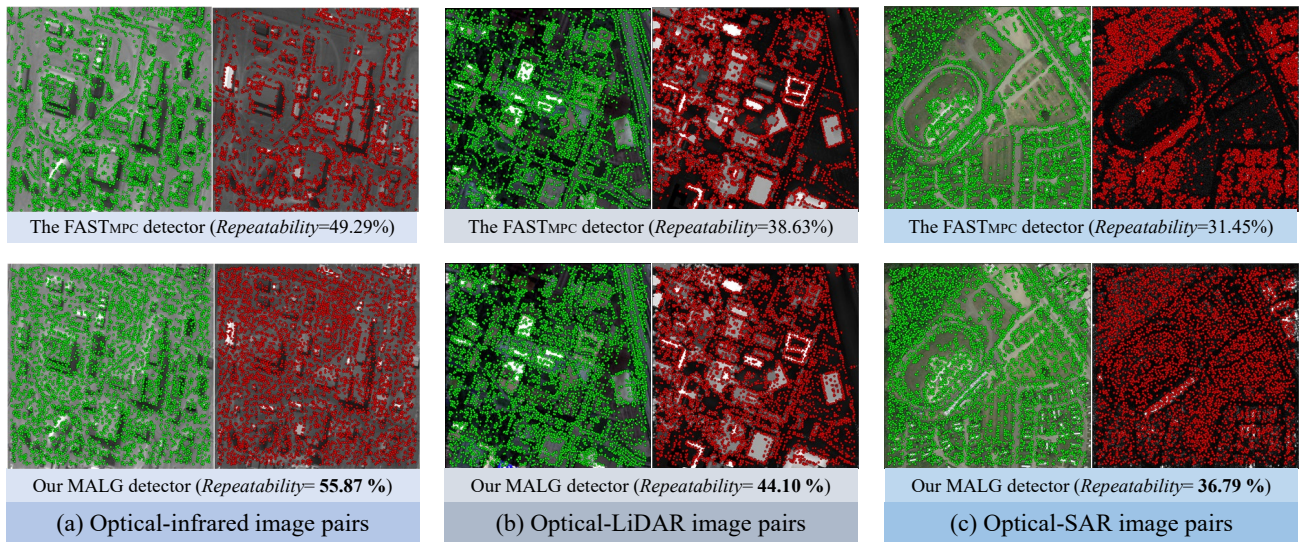

Fig. 8. Comparison of the repeatability and distributions of IPs between the $FAST_{MPC}$ and our MALG detector. (a) is a contrastive example of extracted IPs for optical-infrared image pairs; (b) is a contrastive example of extracted IPs for optical-LiDAR image pairs; (c) is a contrastive example of extracted IPs for optical-SAR image pairs.

As a consequence, our proposed MALG detector, combining the multi-channel Log-Gabor features and auto-correlation strategy, not only improves the repeatability of IPs but also makes the distribution of IPs more uniform, which can lay a foundation for improving the robustness of subsequent feature matching.

## 3.3 Rotation Invariance evaluation of RMLG descriptor

In Section 2.3, we preliminarily demonstrate that the proposed RMIM is invariant to image rotation. However, the calculation of dominant orientation and RMIM both depend on the six orientations (i.e., 0, $\frac{\pi}{6}$, $\frac{2\pi}{6}$, $\frac{3\pi}{6}$, $\frac{4\pi}{6}$, $\frac{5\pi}{6}$) of the log-Gabor filters, so if the rotation angle of image pair is not in the vicinity of the six orientations, can RMIM still maintain its rotation invariance.

In response to this issue, in this section, we further verified the rotation invariance of RMLG. The specific verification process is as follows. First, we randomly selected two sets of image pairs without rotation from the above MRSIs datasets for experimentation. Then, one of the selected image pairs was rotated from 0° to 360° with an interval of 10°, and a total of 72 rotated images can be obtained. These rotated images and their corresponding optical images constitute 72 pairs of test cases, which are finally matched by our $R_2FD_2$. And the two examples of rotation invariance tests for our $R_2FD_2$ are demonstrated in the following Fig. 9.

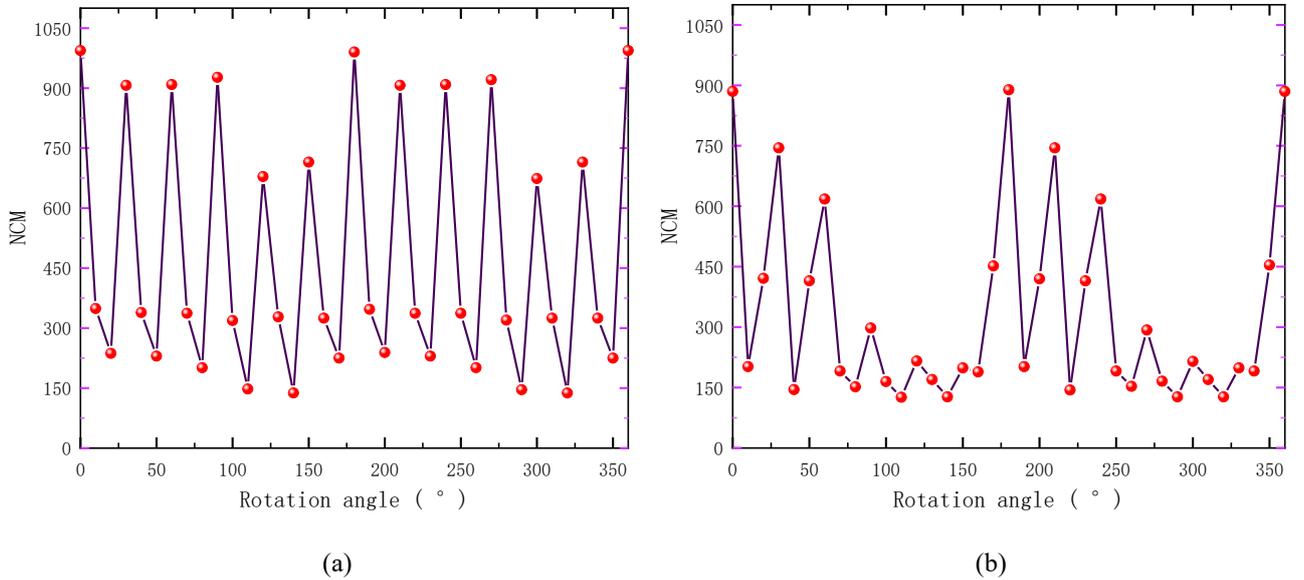

Fig. 9. Rotation invariance test of $R_2FD_2$ on different types of MRSIs datasets as the rotation angles from 0° to 360°. (a) NCMs of first tested image pairs. (b) NCMs of second tested image pairs.

These NCMs were marked with red dots, it can be clearly seen that NCMs of all rotated angles were not less than 100, and more than half of NCMs were greater than 300. What's more, the matching success rate of all rotation angles was up to 100%, which further verifies that our $R_2FD_2$ can maintain rotation invariance in the range of [0, 360°]. Fig. 10 shows the matching results of several groups of rotation angles (30°, 130°, 240°, and 350°) and their corresponding registration results. As can be seen, the distribution of correspondences is relatively uniform, and the checkboard maps of registration results have been aligned correctly.

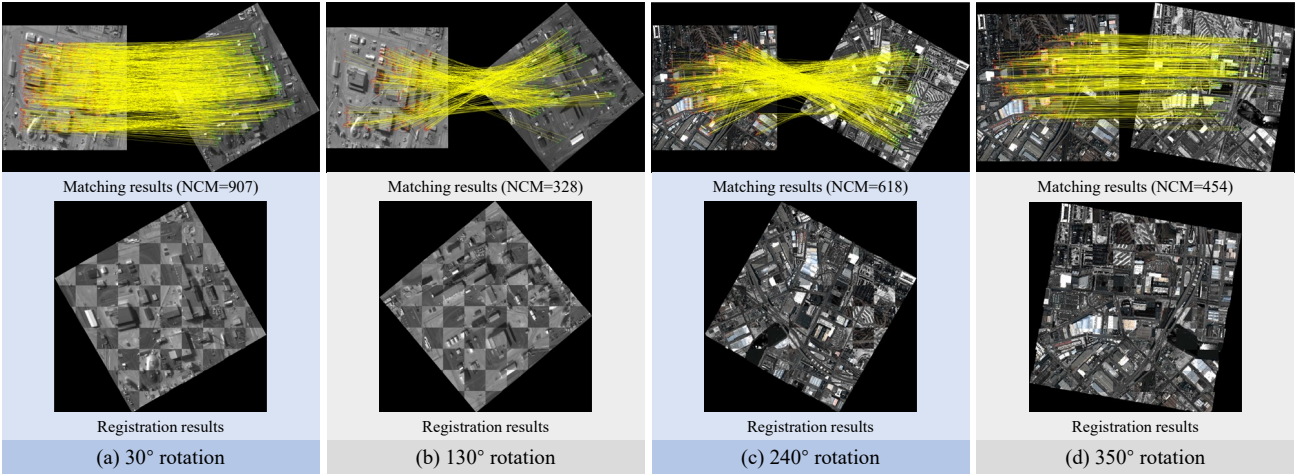

Fig. 10. Visualization of matching and registration results. (a) the matching and registration results of 30° rotation. (b) the matching and registration results of 130° rotation. (c) the matching and registration results of 240° rotation. (d) the matching and registration results of 350° rotation.

### 3.4 Matching performance evaluation of $R_2FD_2$

In this section, we compare our $R_2FD_2$ with the five state-of-the-art matching methods: OS-SIFT, HOSS, RIFT, RI-ALGH, and MS-HLMO. Each image pair of the above MRSIs datasets was rotated from 0° to 180° with an interval of 10°, and a total of 1140 (19*20*3) rotated images can be obtained as experimentation. The correct matches of each image pair were manually determined by selecting 10 to 20 evenly distributed correspondences to estimate the projective model (denoted as $P_{truth}$). These matched correspondences with residuals less than 3 pixels were considered as the correct matches by utilizing the estimated projective model $P_{truth}$. For quantitative evaluation, we employ four criteria to evaluate the performance of each matching method in terms of NCM, Success Rate (SR), Root-Mean-Square Errors (RMSE), and Running time (RT). Among them, NCM represented the number of correspondences correctly matched. If the number of NCM was less than 10, the corresponding image pairs were marked as a matching failure. SR was the ratio between the number of image pairs that are successfully matched to the total number of image pairs. The larger the value of NCM and SR, the stronger the robustness of the corresponding matching method. The RMSE can be calculated as follow:

$$RMSE = \sqrt{\frac{\sum_{i=1}^{N}[R(x,y) - P_{truth} * S(x,y)]^2}{N}} \quad (19)$$

Where $R(x,y)$ and $S(x,y)$ represent the correct matches of the reference and sensed images, respectively. $P_{truth}$ is the projective model, and $N$ represents the number of NCM. The smaller the RMSE, the higher the accuracy of the matching method

As shown in Fig. 11, the comparison results of Average NCM criteria for different matching methods on each multimodal image dataset are demonstrated. And the average NCM refers to the average of all NCMs of a total of 19 sets of images generated by each image pair with an interval of 10 degrees from 0 to 180 degrees. As can be seen, OS-SIFT matched the least NCMs for all types of multimodal image pairs, followed by MS-HLMO. This may be related to the fact that both OS-SIFT and MS-HLMO utilize the Harris-based function to detect IPs, which usually results in the extracted IPs being less than others such as the MALG and $FAST_{MPC}$ detector. The average NCM criteria of HOSS and RIFT were comparable in optical-infrared and optical-LiDAR datasets, while the average NCMs of HOSS were extremely reduced in optical-SAR dataset, RIFT achieved more NCMs than HOSS. This could be caused by the relatively low discriminative capability of HOSS based on LSS description led to the inability to maintain robust matching performance for optical-SAR dataset with significant NRD. In contrast, it is obvious that our $R_2FD_2$ outperformed the other methods in the average NCM criteria, and obtained the most matching numbers on all types of multimodal image pairs, followed by RI-ALGH. This indicates that the features detected by our MALG are more repeatable and the features described by our RMLG are more discriminative.

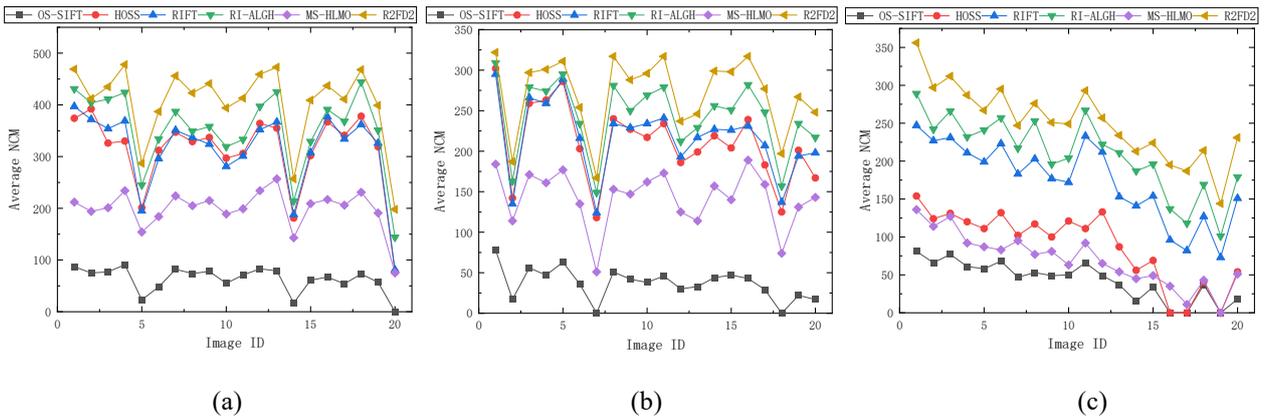

Fig. 11. Comparisons of Average NCM criteria for different matching methods. (a) Average NCM of optical-infrared datasets. (b) Average NCM of optical-LiDAR datasets. (c) Average NCM of optical-SAR datasets.

Fig. 12 depicts the comparison results of SR criteria for different matching methods, where the SR criteria of OS-SIFT was the worst for optical-infrared and optical-SAR datasets, especially since there are cases where the SR of OS-SIFT was zero in each type of dataset. HOSS, RIFT, RI-ALGH, and MS-HLMO had comparable performance regarding the SR criteria in optical-infrared datasets, while HOSS achieved the lowest SR for optical-SAR datasets. On the whole, our $R_2FD_2$ obtained the highest SR on all the datasets, the matching success rate of $R_2FD_2$ reached 100% on most datasets and closed to 100% on a few image pairs.

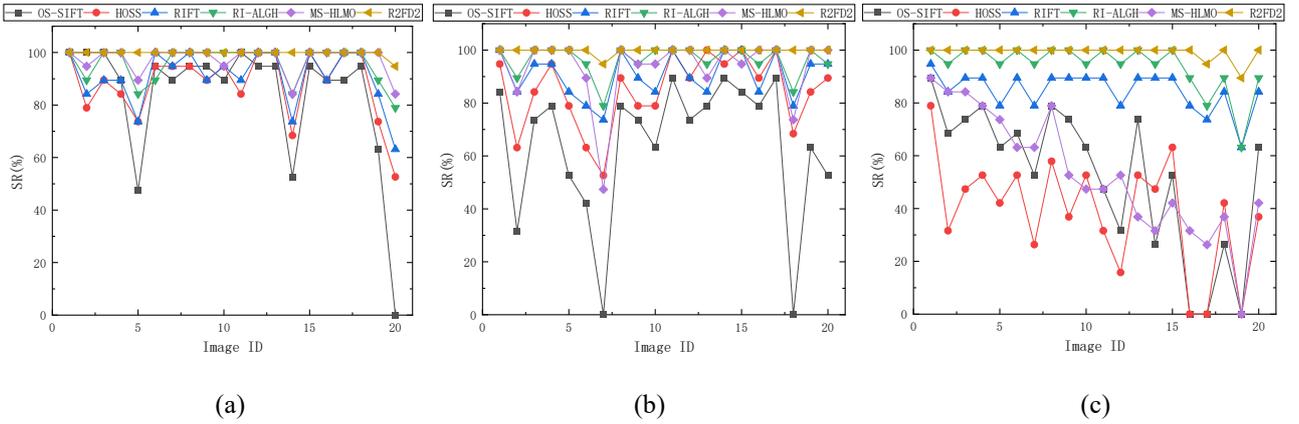

Fig. 12. Comparisons of SR criteria for different matching methods. (a) SR of optical-infrared datasets. (b) SR of optical-LiDAR datasets. (c) SR of optical-SAR datasets.

To further evaluate the accuracy of different matching methods, Fig. 13 shows the comparison results of Average RMSE criteria, where RMSE was set to 10 indicating a failed match. Similar to the average NCM, the average RMSE refers to the average of all RMSEs of a total of 19 sets of images generated by each image pair with an interval of 10 degrees from 0 to 180 degrees. It can be seen from Fig. 13 that our $R_2FD_2$ yielded the best results on the criterion of average RMSE and achieved the matching accuracy of fewer than 2 pixels for all datasets. This was followed by RI-ALGH and RIFT, their RMSE is relatively worse than our $R_2FD_2$. Nevertheless, OS-SIFT, HOSS, and MS-HLMO were all likely to exhibit the worst performance on the criterion of average RMSE for different cases. These experimental results further illustrate that the validity of the proposed approaches as compared to the state-of-the-art methods, and the rotation-invariance performed by the RMLG descriptor is more reliable than others.

Table 3 gives the average RT of each compared method for the whole dataset, which was implemented on a laptop with a CPU i7-10750H 2.6 GHz and 16 GB RAM. The average RT of our $R_2FD_2$ was the fastest, and the time consumption was about 11 seconds. The efficiency of OS-SIFT was second best, RIFT ranked third, HOSS and MS-HLMO fourth, and RI-ALGH last. Specifically, the RT of our $R_2FD_2$ was about 9 times, 6.5 times, 5 times, and 2 times faster than RI-ALGH, MS-HLMO (HOSS), RIFT, and OS-SIFT, respectively. It is obvious that our $R_2FD_2$ has a great advantage in matching efficiency, which is attributed to the fast assignment of dominant orientation utilizing and the construction of the RMLG descriptor using the RMLG with rotation invariance.

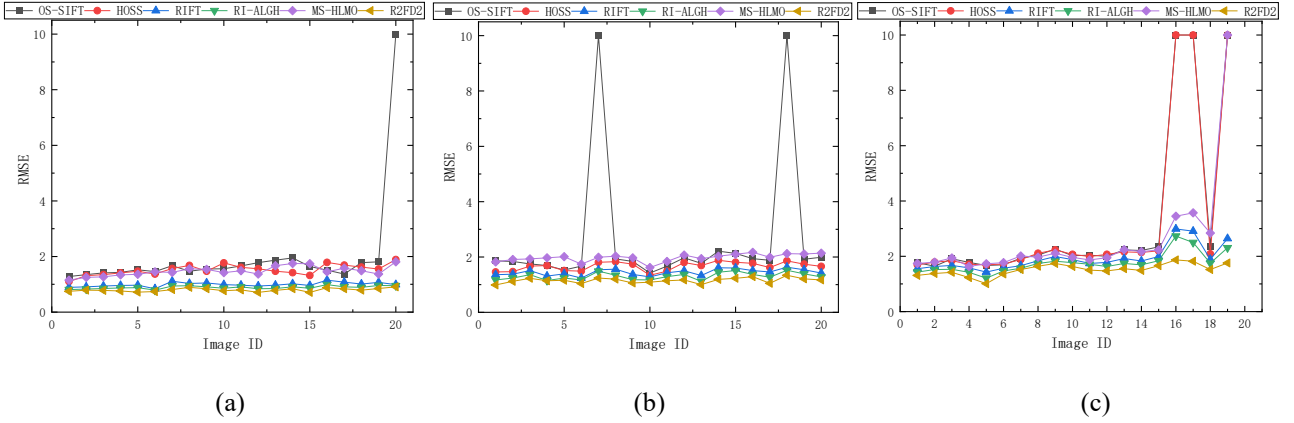

Fig. 13. Comparisons of Average RMSE criteria for different matching methods. (a) RMSE of optical-infrared datasets. (b) RMSE of optical-LiDAR datasets. (c) RMSE of optical-SAR datasets.

Table 3. Comparisons of RT criteria for each matching method.

| Method | OS-SIFT | HOSS | RIFT | RI-ALGH | MS-HLMO | $R_2FD_2$ |
|--------|---------|-------|-------|---------|---------|-----------|
| RT     | 22.57   | 69.21 | 58.41 | 97.94   | 73.69   | 11.27     |

Furthermore, we carried out qualitative evaluations for $R_2FD_2$ by displaying the correct correspondences and registration results for the visual inspection, respectively. Fig. 14 shows more matching results of our $R_2FD_2$, and at least four image pairs were randomly selected from each multi-modal dataset and applied different rotation deformations in the range [0, 360°]. Fig. 15 displays the corresponding registration results of Fig. 14 by using checkboard maps. Each edge of all chessboard maps can be well aligned without obvious misalignment, which further verifies the satisfactory generality of $R_2FD_2$.

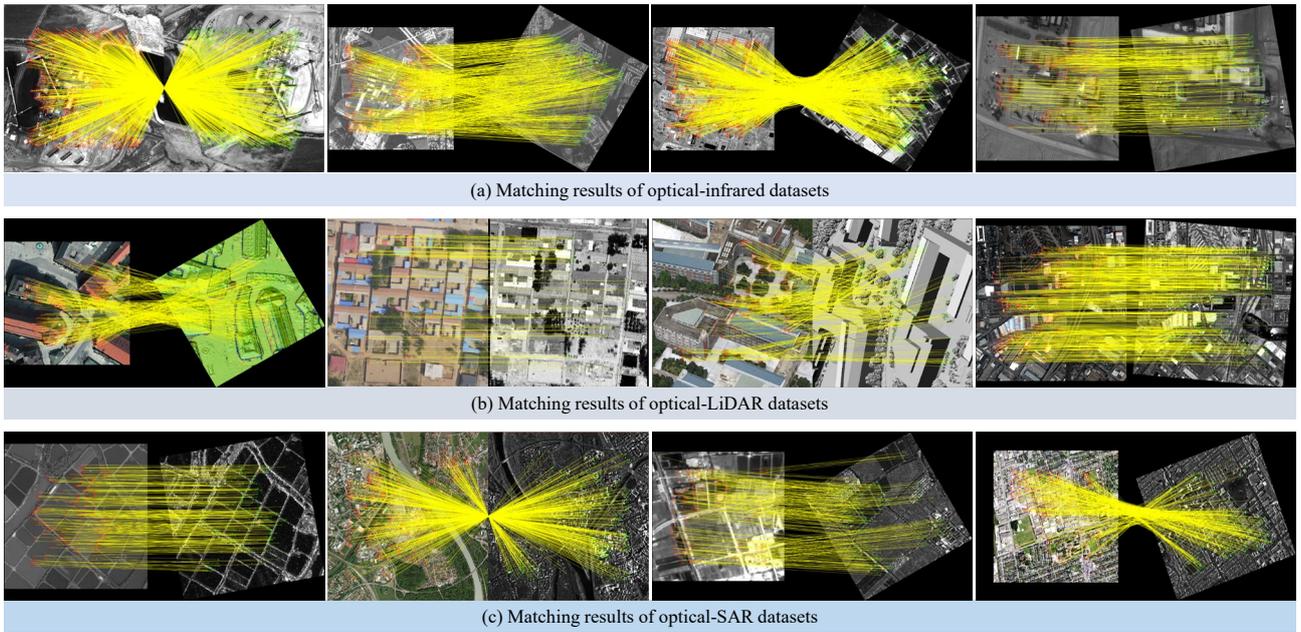

(a) Matching results of optical-infrared datasets

(b) Matching results of optical-LiDAR datasets

(c) Matching results of optical-SAR datasets

Fig. 14. Correspondence visualization of $R_2FD_2$. (a) matching results of optical-infrared datasets. (b) matching results of optical-LiDAR datasets. (c) matching results of optical-SAR datasets.

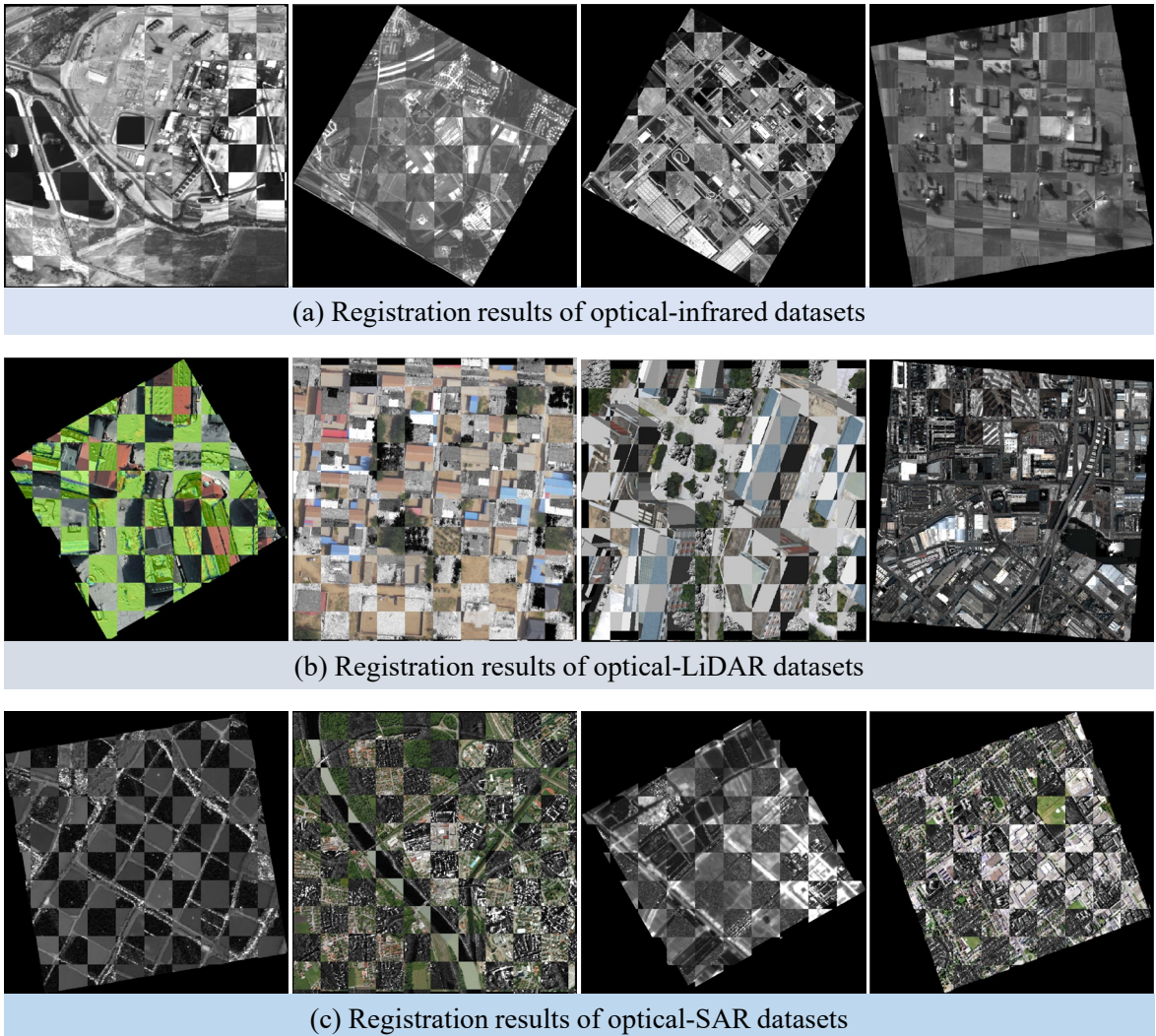

(a) Registration results of optical-infrared datasets

(b) Registration results of optical-LiDAR datasets

(c) Registration results of optical-SAR datasets

Fig. 15. Checkboard visualization of $R_2FD_2$. (a) registration results of optical-infrared datasets. (b) registration results of optical-LiDAR datasets. (c) registration results of optical-SAR datasets.

Overall, these evaluations and coherence analysis proved that our $R_2FD_2$ achieved high computational efficiency and the effectiveness of our $R_2FD_2$ in resisting significant radiation and rotation differences among multimodal images were far superior to the state-of-the-art feature matching methods. The excellent matching performance of $R_2FD_2$ was mainly due to the following two reasons. On the one hand, the feature detection of $R_2FD_2$ adopted the novel MALG detector, and MALG had the excellent property of high repeatability and uniform distribution for IPs detection, which can be rather advantageous for subsequent matching. On the other hand, the feature description of $R_2FD_2$ utilized the discriminative RMLG descriptor, RMLG integrated the rotation-invariant RMIM with the arrangement of DAISY to produce locally invariant features, which lays a foundation for fast and robust matching.

## 4. CONCLUSION

In this paper, a novel feature matching method (named $R_2FD_2$) was presented for MRSIM, involving both the repeatable MALG detector and the rotation-invariant RMLG descriptor. MALG detector was first designed by integrating the multi-channel auto-correlation strategy with the Log-Gabor wavelets for IPs extraction. In this way, IPs extracted by MALG generally had a high repetition rate and were evenly distributed in multimodal images. Then, the fast assignment strategy of dominant orientation was proposed to establish the novel RMIM with rotation invariance. Subsequently, RMLG descriptor was conducted by incorporating the rotation-invariant RMIM with the spatial arrangement of DAISY for feature representation. Qualitative and quantitative experiments were performed by utilizing different types of MRSIs datasets (optical-infrared, optical-LiDAR, and optical-SAR image pairs) to evaluate the matching performance of our $R_2FD_2$. The experimental results demonstrated that the proposed $R_2FD_2$ outperformed five state-of-the-art feature matching methods (i.e., OS-SIFT, HOSS, RIFT, RI-ALGH, and MS-HLM) in all criteria (including NCM, SR, RMSE, and RT). As a result, our $R_2FD_2$ can be capable of reliably achieving fast and robust feature matching for MRSIs.

Although the proposed $R_2FD_2$ exhibited superior adaptation to rotation and radiation differences for multimodal feature matching, it was sensitive to scale distortions between multimodal images because it did not address the question of scale invariance. Accordingly, our future research will include the exploration of these limitations more deeply. For example, it is of great significance to establish a suitable scale space for achieving scale invariance, such as co-occurrence scale space (Yao et al., 2022), nonlinear diffusion scale space (Fan et al., 2018), and Gaussian scale space (Dellinger et al., 2014).

## 5. ACKNOWLEGEMENTS

The funding for the research discussed in this paper was provided by the National Natural Science Foundation of China (No. 42271446 and No. 41971281).